\documentclass[conference]{IEEEtran}

\usepackage{times}

\usepackage[numbers]{natbib}
\usepackage{multicol}
\usepackage[hidelinks,draft]{hyperref}
\usepackage{graphicx}

\usepackage{latexsym}
\usepackage{url}
\usepackage{amssymb}
\usepackage{amsmath}
\usepackage{dsfont}
\usepackage{multirow}
\usepackage{tikz,pgfplots}
\usepackage{balance}
\usepackage{booktabs}
\usepackage{tikz}

\usepackage{amsthm}

\newtheorem*{definition*}{Definition}
\newtheorem*{theorem*}{Theorem}

\newcommand{\software}{\mbox{CHALET}}
\newcommand{\nlstring}[1]{{\em #1}}

\title{CHALET: Cornell House Agent Learning Environment}
\author{
\begin{tabular}{ccc}
  \textbf{Claudia Yan}$^{\heartsuit}$$^{\diamondsuit}$ & \textbf{Dipendra Misra}$^{\diamondsuit}$ & \textbf{Andrew Bennett}$^{\diamondsuit}$ \\
  \textbf{Aaron Walsman}$^{\clubsuit}$ & \textbf{Yonatan Bisk}$^{\clubsuit}$ & \textbf{Yoav Artzi}$^{\diamondsuit}$
\end{tabular}
\\[.2cm]
 $^{\diamondsuit}$ Cornell University \,\,$^{\heartsuit}$ City College of New York  \,\,$^{\clubsuit}$ University of Washington\\[.2cm]

$^{\diamondsuit}$\emph{\{dkm, awbennett, yoav\}}@cs.cornell.edu  \,\, $^{\heartsuit}$\emph{cyan000@citymail.cuny.edu} \,\,$^{\clubsuit}$\emph{\{awalsman, ybisk\}}@cs.washington.edu\\[10pt]
{\large \href{https://github.com/lil-lab/chalet}{\textcolor{purple}{https://github.com/lil-lab/chalet}}}\\
}

\begin{document}

\maketitle

\begin{abstract}
We present \software{}, a 3D house simulator with support for navigation and manipulation. \software{} includes 58 rooms and 10 house configuration, and allows to easily create new house and room layouts. \software{} supports a range of common household activities, including moving objects, toggling appliances, and placing objects inside closeable containers. The environment and actions available are designed to create a challenging domain to train and evaluate autonomous agents, including for tasks that combine language, vision, and planning in a dynamic environment. 
\end{abstract}

\IEEEpeerreviewmaketitle

\section{Introduction}

Training autonomous agents poses challenges that go beyond the common use of annotated data in supervised learning. 
The large set of states an agent may observe and the importance of agent behavior in identifying states for learning require  interactive training environments, where the agent observes the outcome of its behavior and receives feedback. 
While physical environments easily satisfy these requirements, they are costly, difficult to replicate, hard to scale, and require complex robotic agents. 
These challenges are further exacerbated by the increased focus on neural network policies for agent behavior, which require significant amounts of training data~\cite{Mnih:13atari,salimans2017evolution}. 
Recently, these challenges are addressed with simulated environments~\cite{Mnih:13atari,Misra:17instructions,beattie2016deepmind,Zhu:16TargetdrivenVN,
Kempka16,Brockman:openaigym,johnson2016malmo,Wu:17,Savva:17,Anderson:17,Brodeur:17}.
In this report, we introduce the Cornell House Agent Learning EnvironmenT (\software{}), an interactive house environment. 
\software{} supports navigation and manipulation of both objects and the environment. It is implemented using the Unity game development engine, and can be deployed on various platforms, including web environments for crowdsourcing. 

\newcommand{\tablefigwide}[1]{ \frame{\includegraphics[width=0.29\textwidth]{{#1}}}} 
\begin{figure*}[!htb]
\centering

  \fbox{\begin{minipage}{0.98\linewidth}
  \centering
  \vspace{5pt}
  \begin{tabular}{ccc}
    \tablefigwide{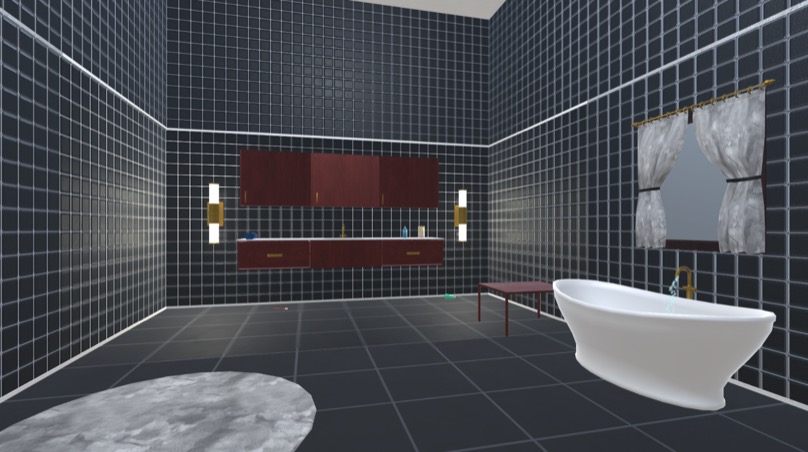} & 
    \tablefigwide{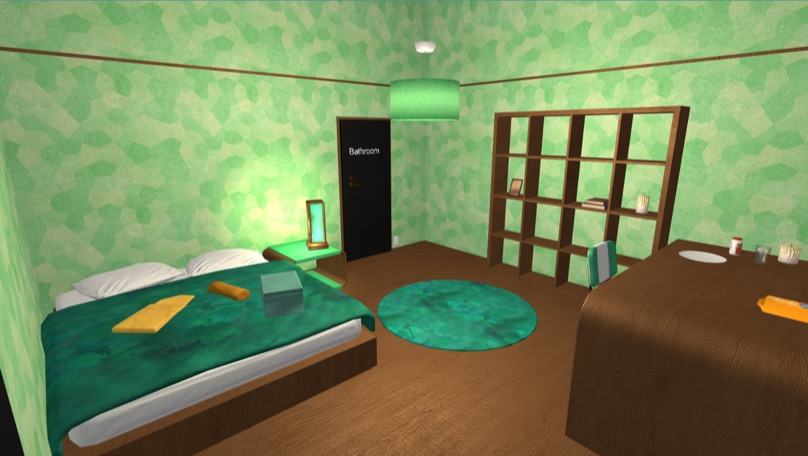} & 
    \tablefigwide{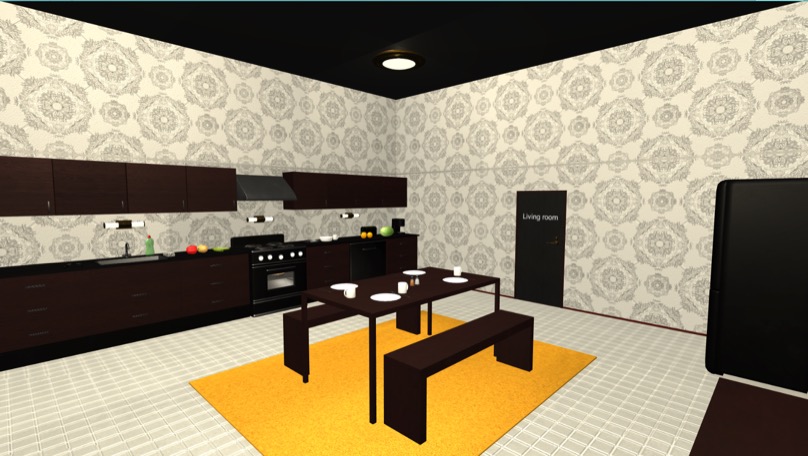} \\
    \tablefigwide{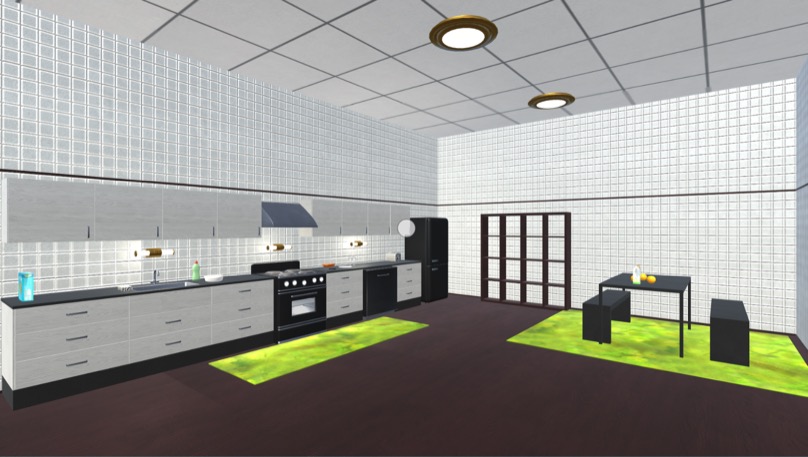} & 
    \tablefigwide{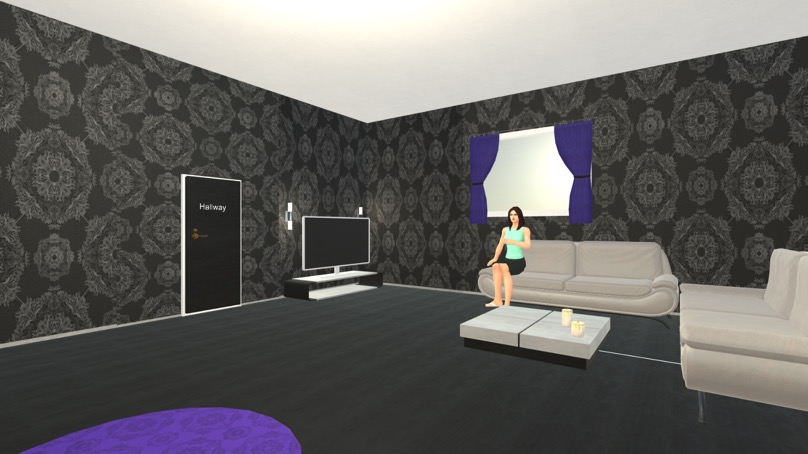} & 
    \tablefigwide{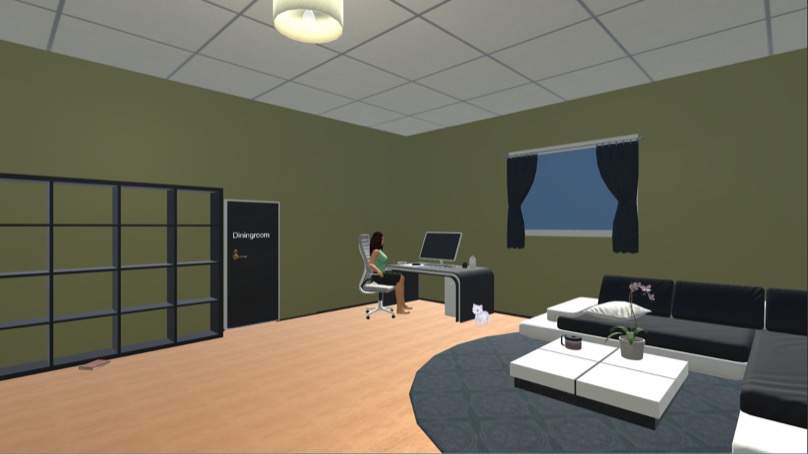} \\
    \tablefigwide{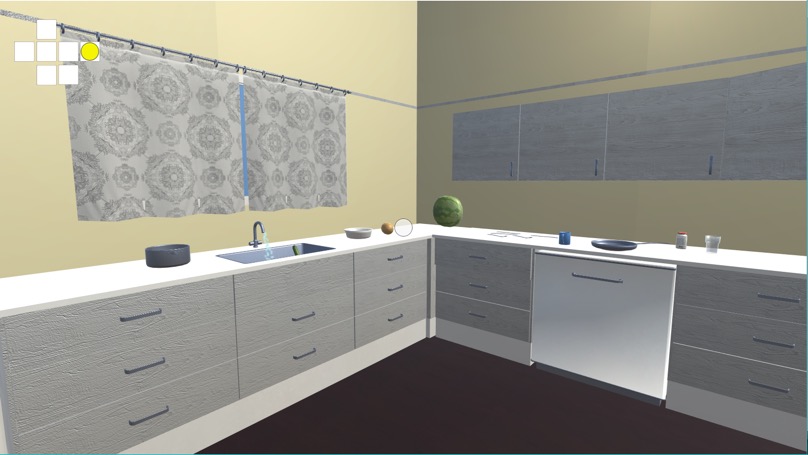} & 
    \tablefigwide{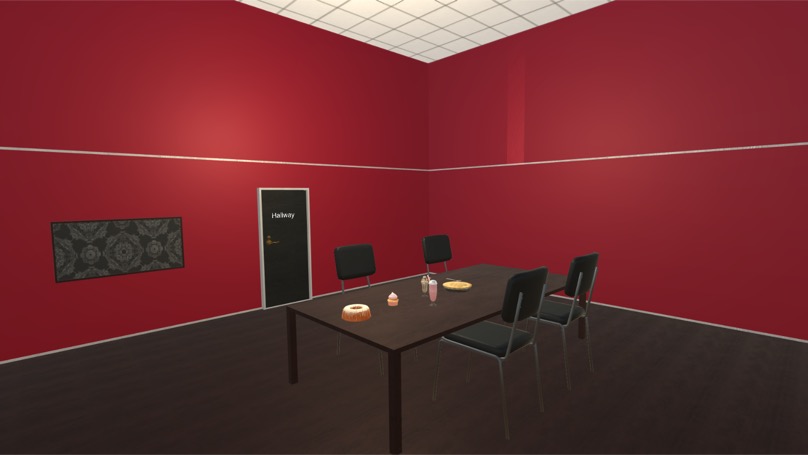} & 
    \tablefigwide{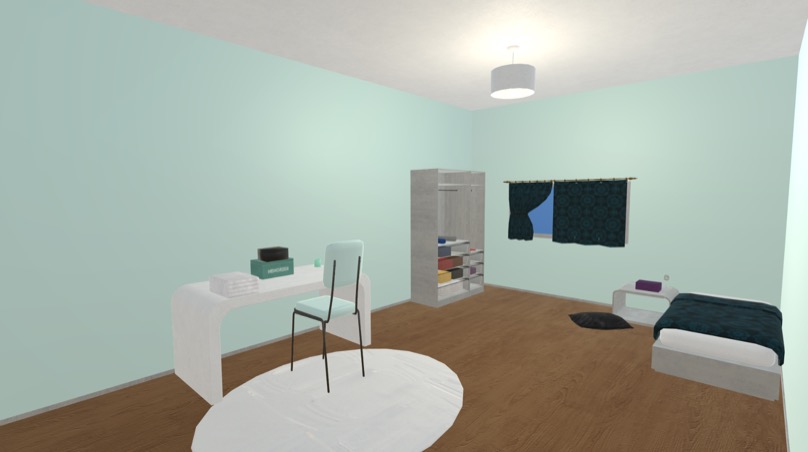} \\
    \tablefigwide{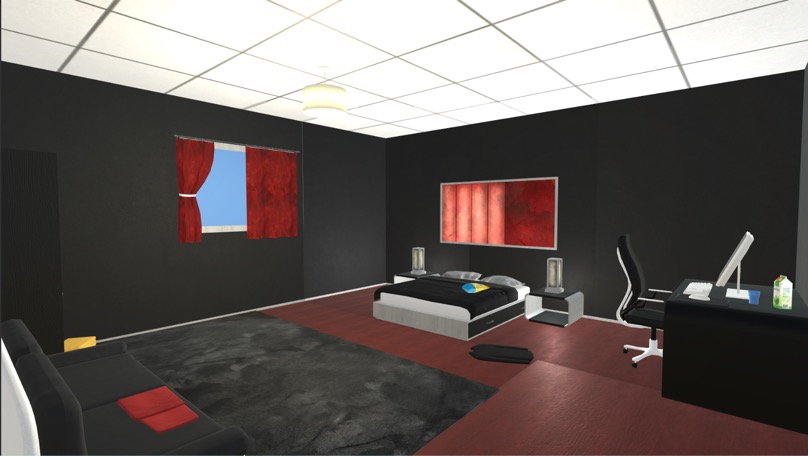} & 
    \tablefigwide{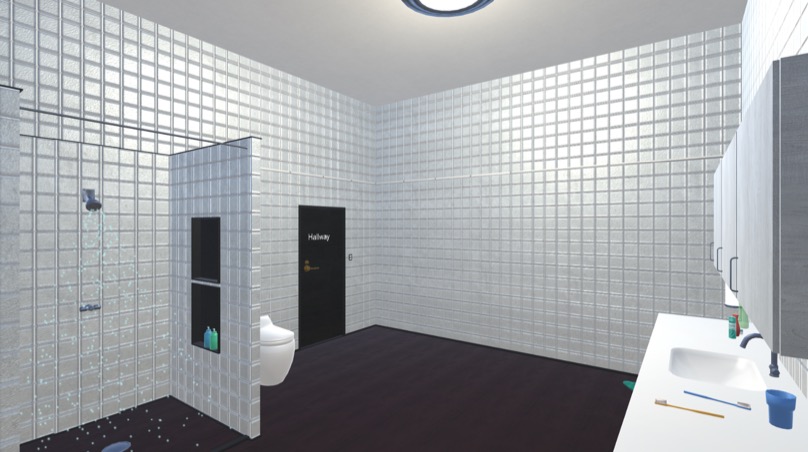} & 
    \tablefigwide{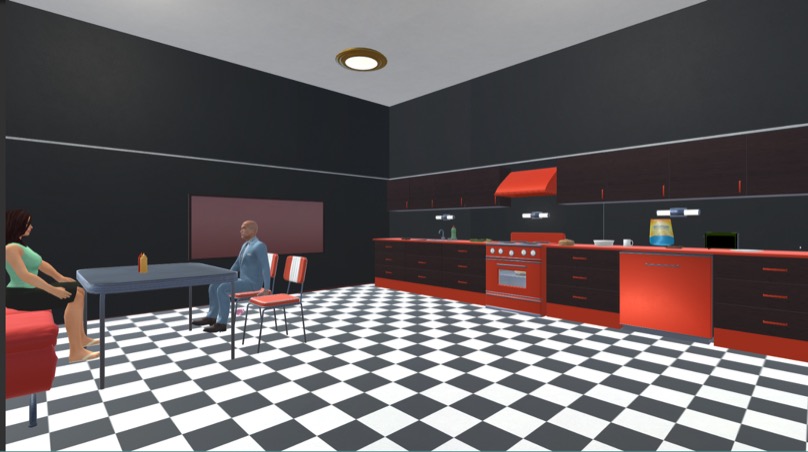} 
  \end{tabular}
  \vspace{5pt}
  \end{minipage}}\\[1.5em]

  \caption{Screenshots of various rooms from \software{}. Each house includes 4-7 rooms of various kinds, including bathrooms, bedrooms, and kitchens.} 
  \label{fig:sample}

\end{figure*}

\section{Environment}

\software{} includes 58 rooms organized into 10 houses. Figure~\ref{fig:sample} shows a sample of the rooms.
The environment contains 150 object types (e.g., fridge, sofa, plate). 
71 types of objects can be manipulated: 60 picked and placed (e.g., plates and towels), 6 opened and closed (e.g., dishwashers and cabinets), and 5 change their state (e.g., opening or closing a faucet). 
Object types are used with different textures to generate 330 different objects. 
On average, each room includes 30 objects. 
Rooms often contain multiple objects of the same kind. For example, kitchens contain many plates and glasses, and bathrooms contain multiple towels. 
Objects that can be opened and closed are container objects, and can contain other objects. 
For example, opening a dishwasher exposes a set of racks, and pulling a rack out allows the agent access to the objects on that rack. The agent can also put an object on the rack, close the dishwasher, and open it later to retrieve the object.  Figure~\ref{fig:continuous-control} shows example object manipulations. 
The environment supports simple physics, including collision-detection and gravity. 

\begin{table*}
\centering
\begin{tabular}{|l|p{14cm}|}
\hline
\textbf{Action} & \textbf{Description}\\
\hline
move-forward & Change the agent location in the direction of its current orientation \\
move-back & Change the agent location in the direction opposite to its current orientation \\
strafe-right & Change the agent location in the direction of $90^\circ$ to its current orientation \\ 
strafe-left & Change the agent location in the direction of $270^\circ$ to its current orientation \\ 
look-left & Change the agent orientation to left \\
look-right & Change the agent orientation to right \\
look-up & Change the agent orientation up (when engaged with a container, change container towards closure) \\
look-down & Change the agent orientation down (when engaged with a container, change container towards open) \\
interact & Engage the container at the current orientation, pick the object at the current orientation, drop the object currently held, toggle state of object at current orientation (e.g., toggle TV power) \\
\hline
\end{tabular}
\caption{The actions available to the agent in \software{}.}
\label{tbl-action}
\end{table*}

\newcommand{\tablefig}[1]{ \frame{\includegraphics[width=0.14\textwidth]{{#1}}}}

  \begin{figure*}[!h]
         \fbox{\begin{minipage}{0.98\linewidth}
           \centering
          \vspace{5pt}
            \begin{tabular}{cccccc}
                \tablefig{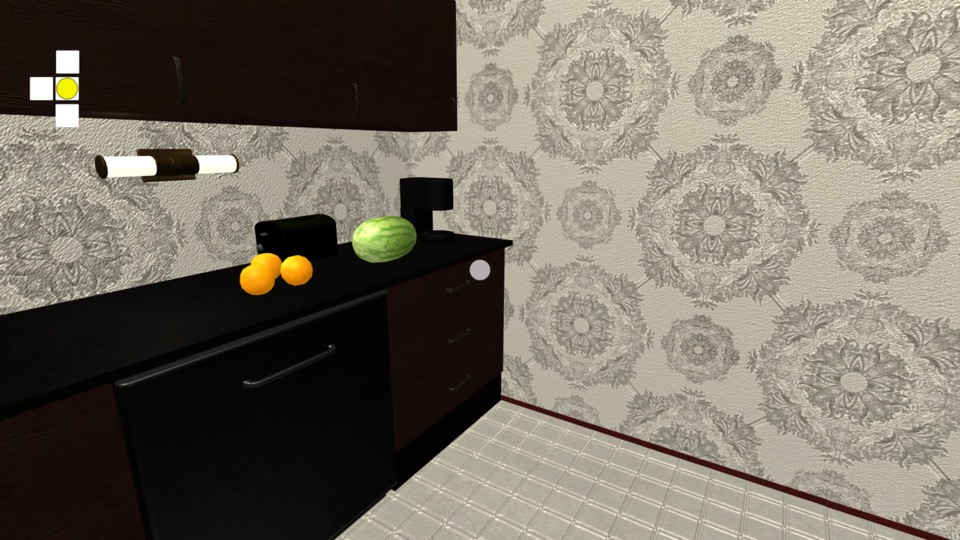} & 
                \tablefig{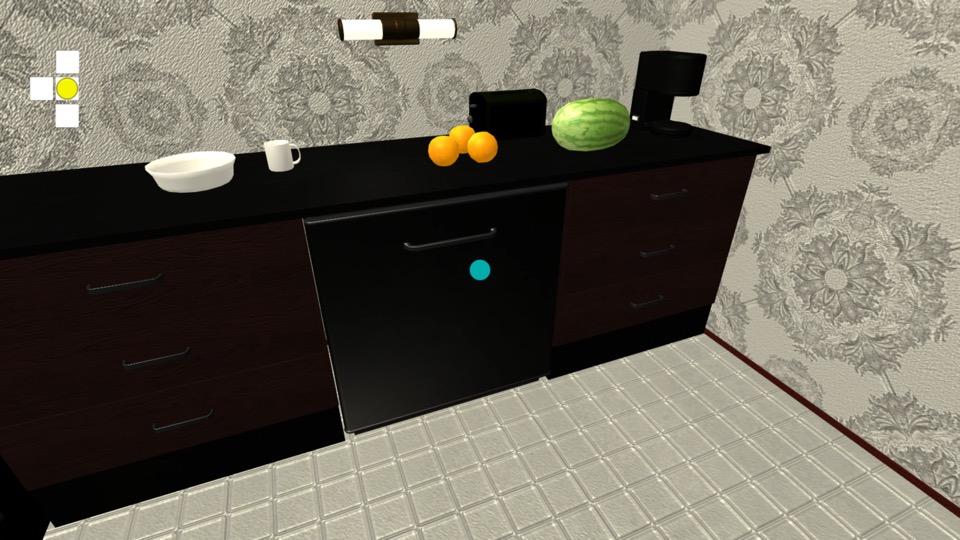} & 
                \tablefig{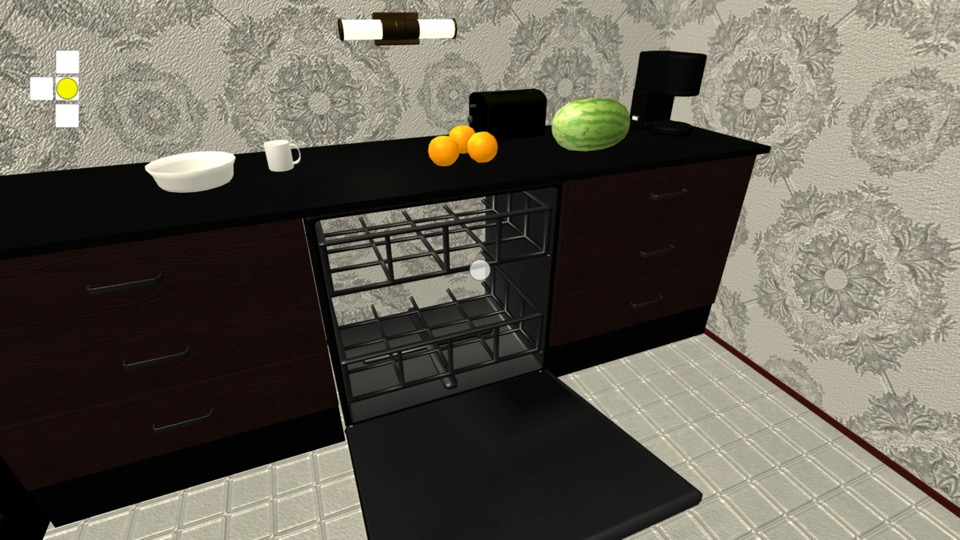} &
                \tablefig{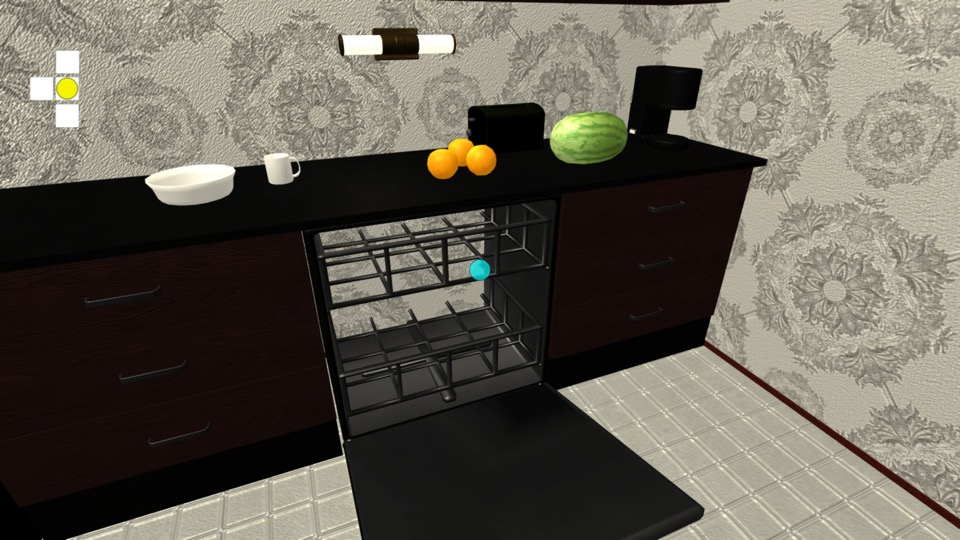} &
                \tablefig{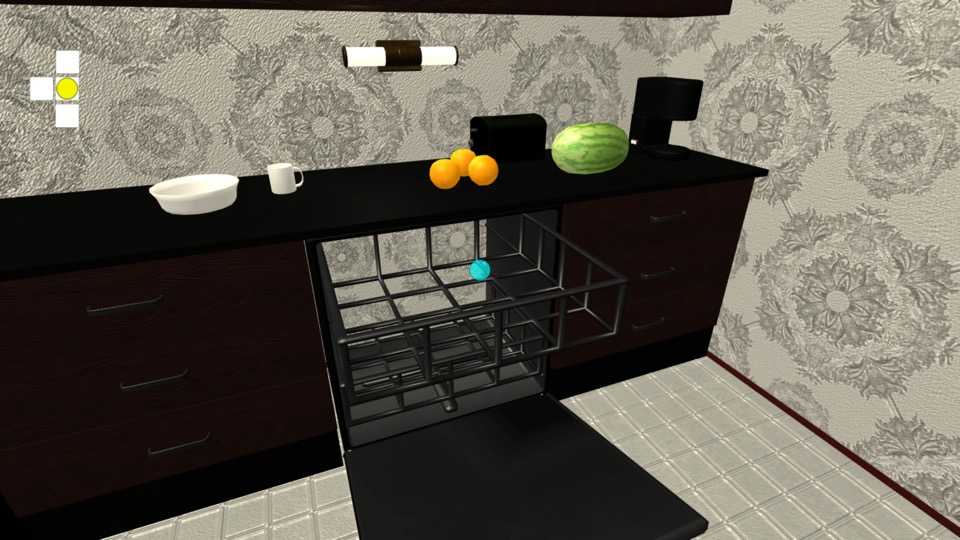} &
                \tablefig{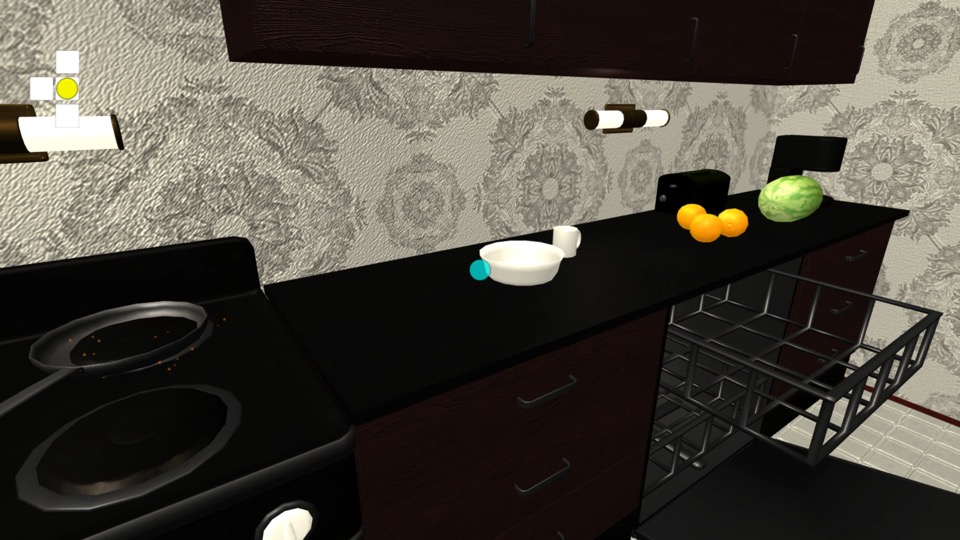} \\
                \tablefig{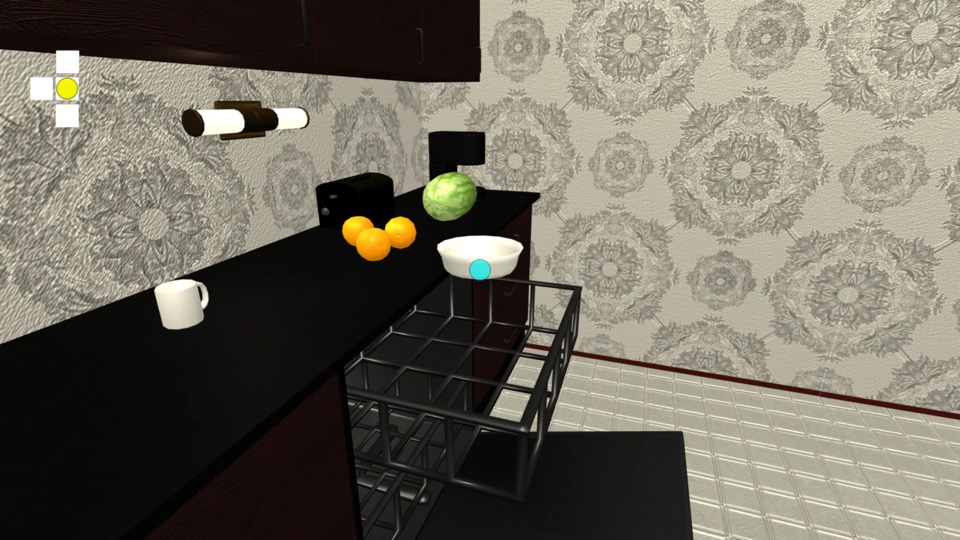} & 
                \tablefig{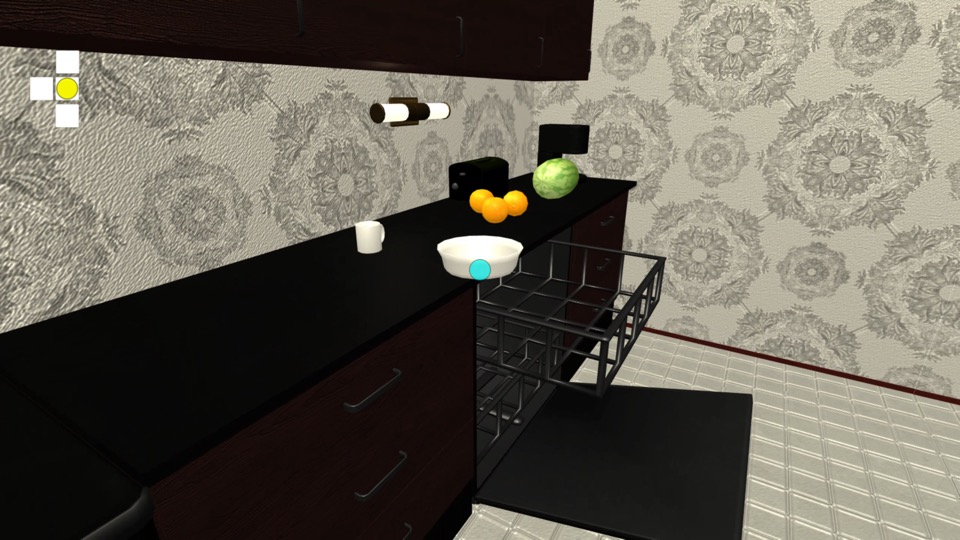} & 
                \tablefig{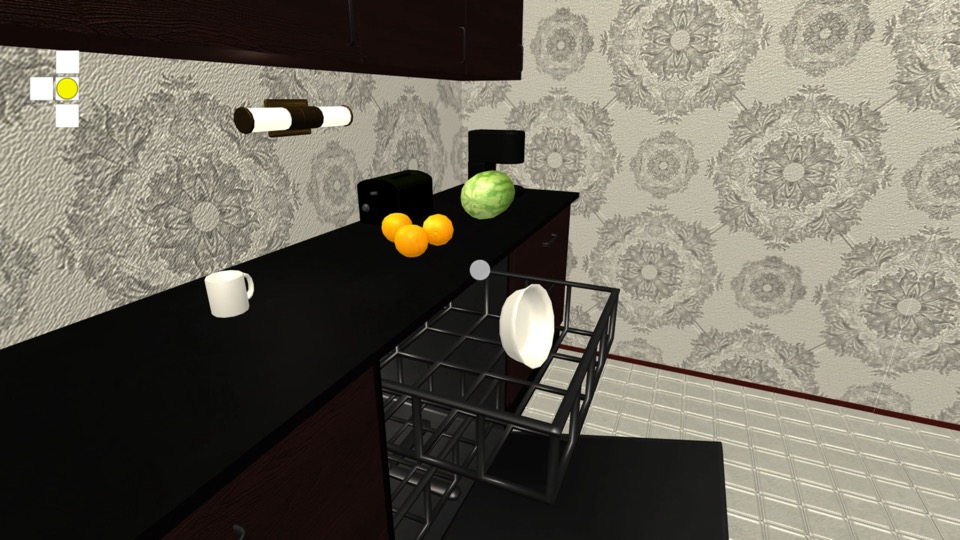} &
                \tablefig{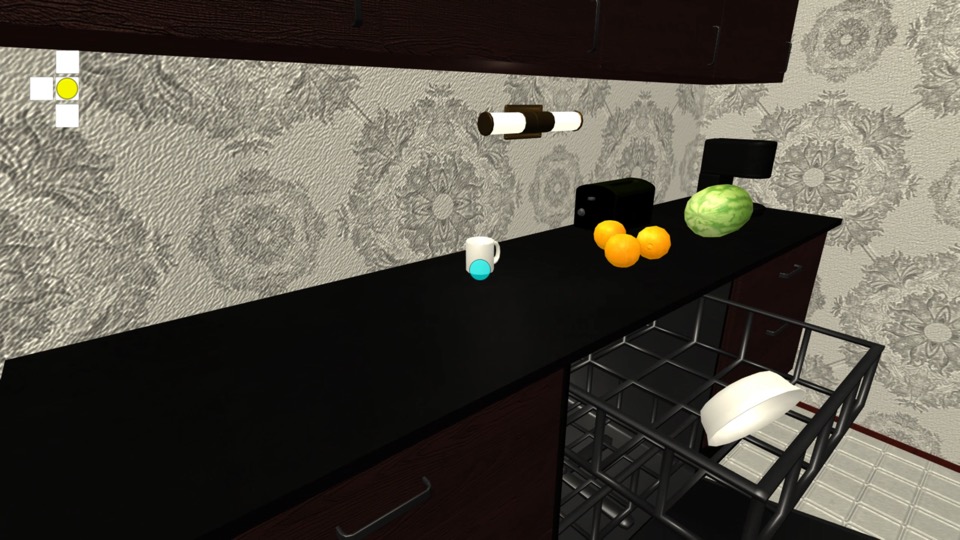} &
                \tablefig{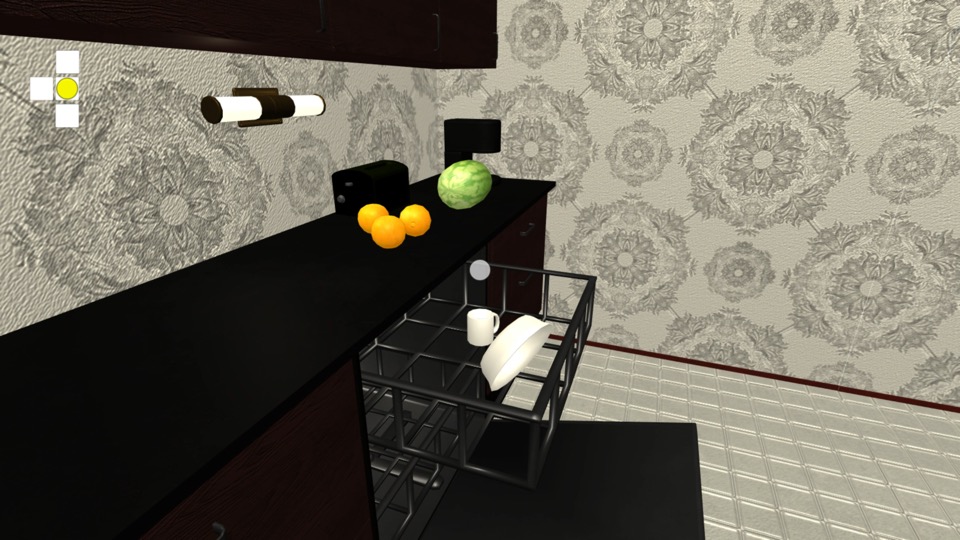} &
                \tablefig{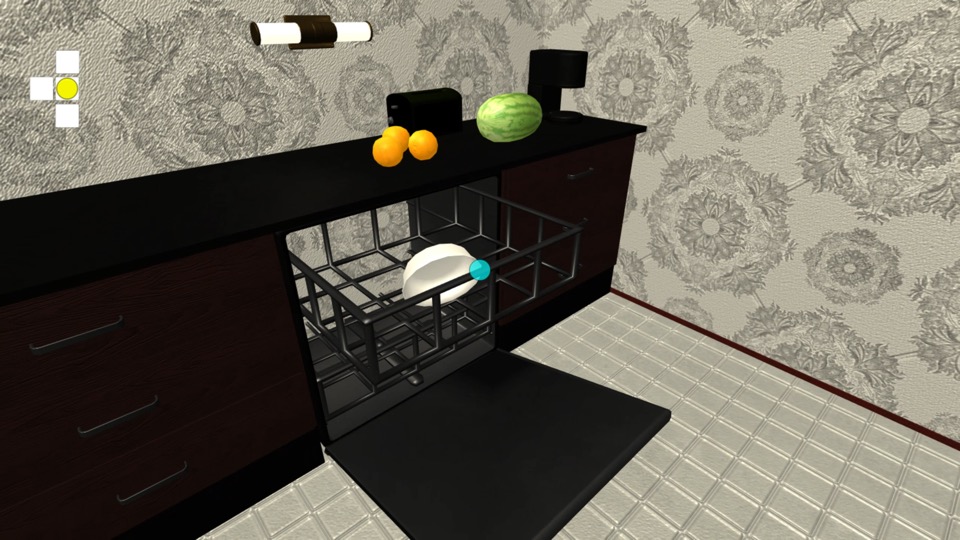}
            \end{tabular}
            \vspace{5pt}
            \end{minipage}}\\[1.5em]
            
      		\fbox{\begin{minipage}{0.98\linewidth}
      		\centering
          \vspace{5pt}
      		\begin{tabular}{cccccc}            
                  \tablefig{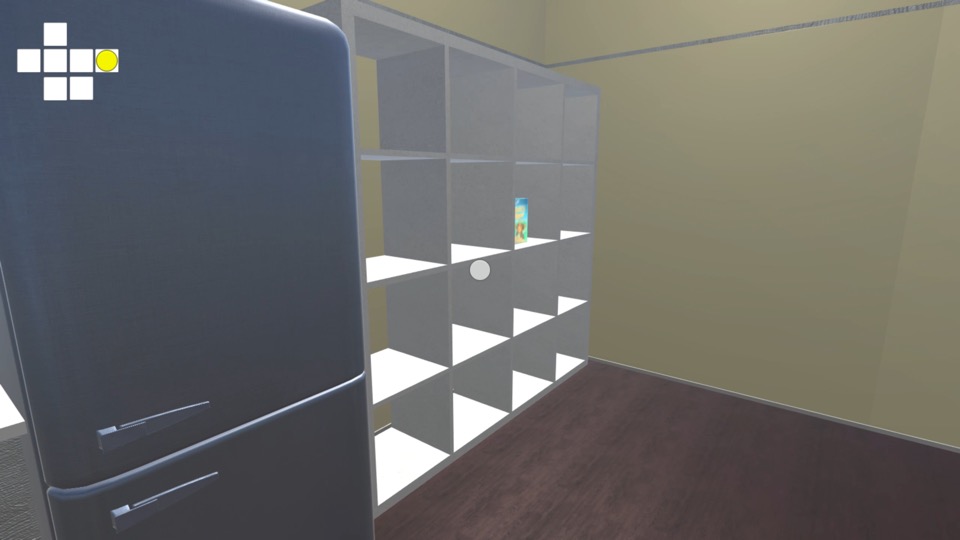} & 
                  \tablefig{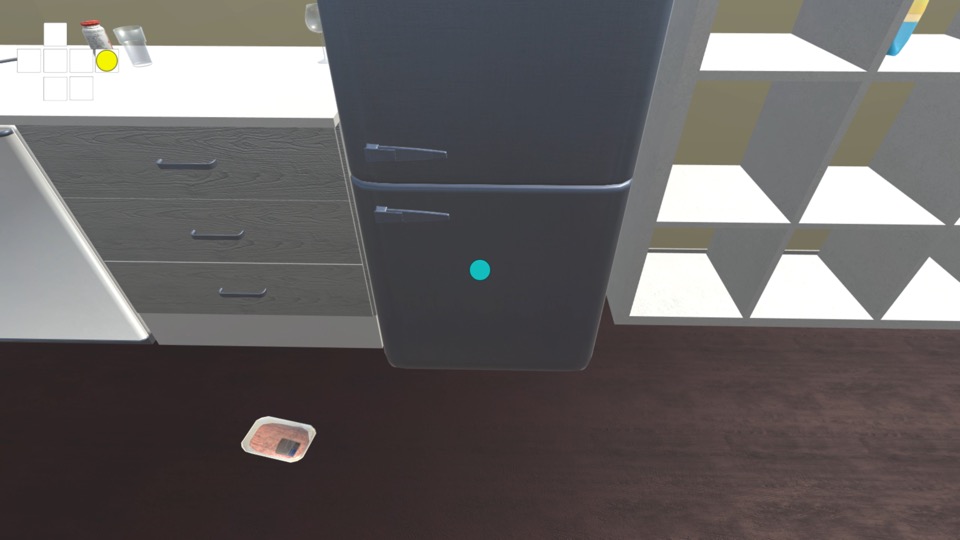} &
                  \tablefig{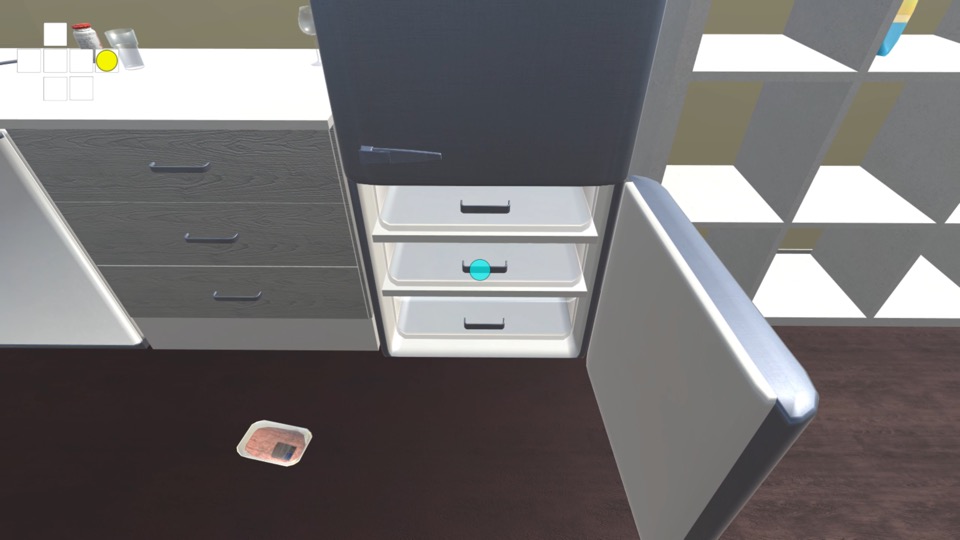} &
                  \tablefig{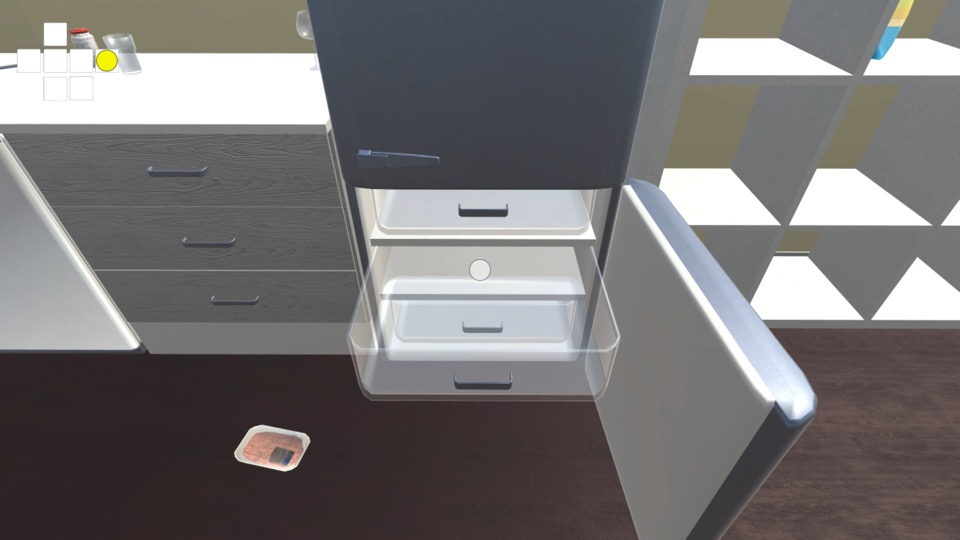} &
                  \tablefig{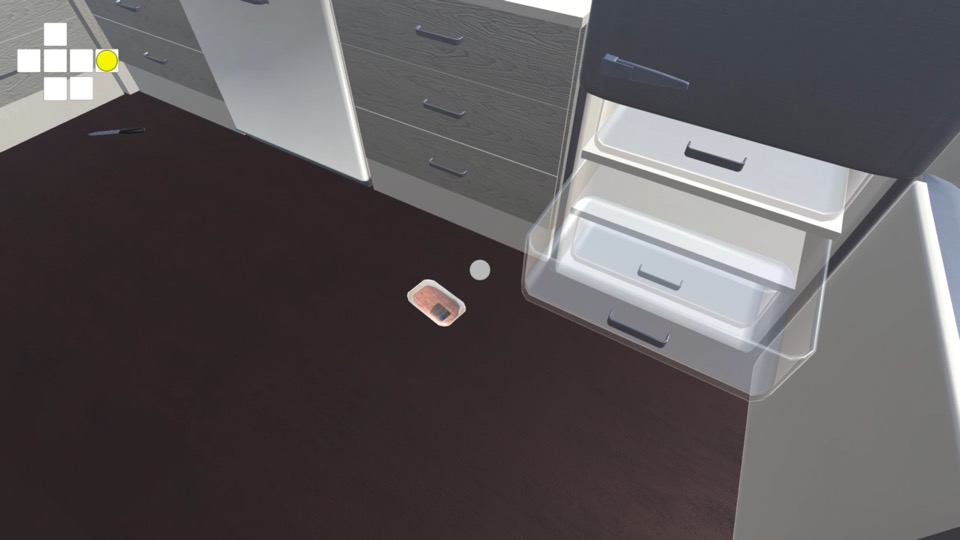} &
                  \tablefig{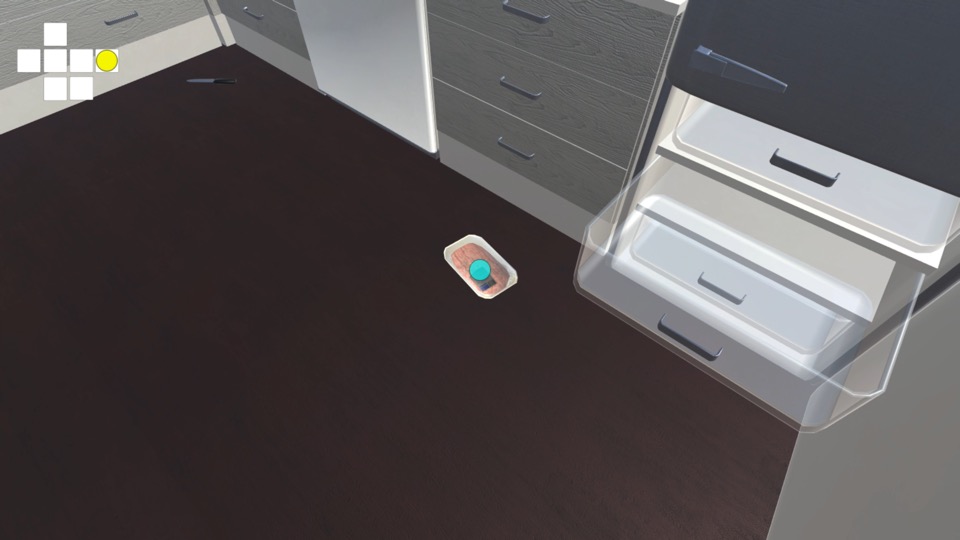} \\
                  \tablefig{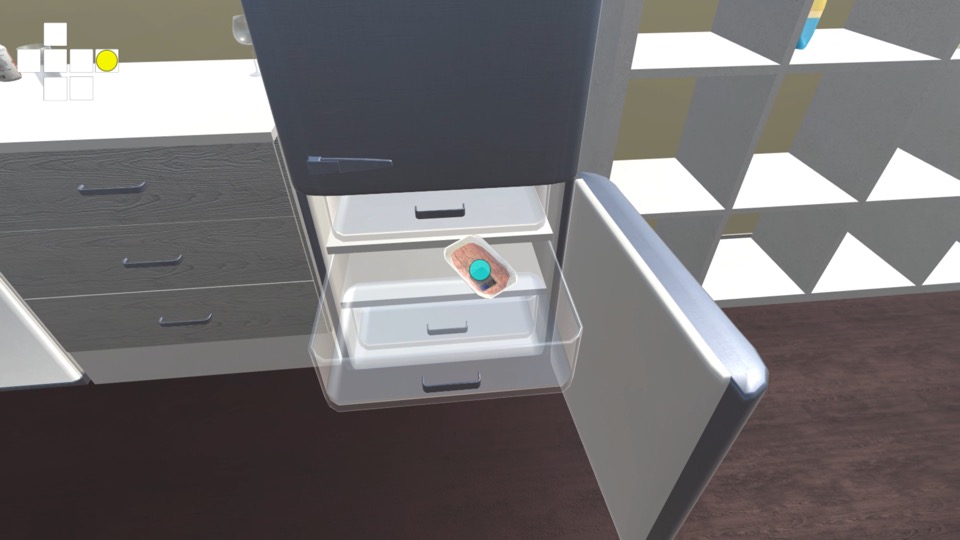} & 
                  \tablefig{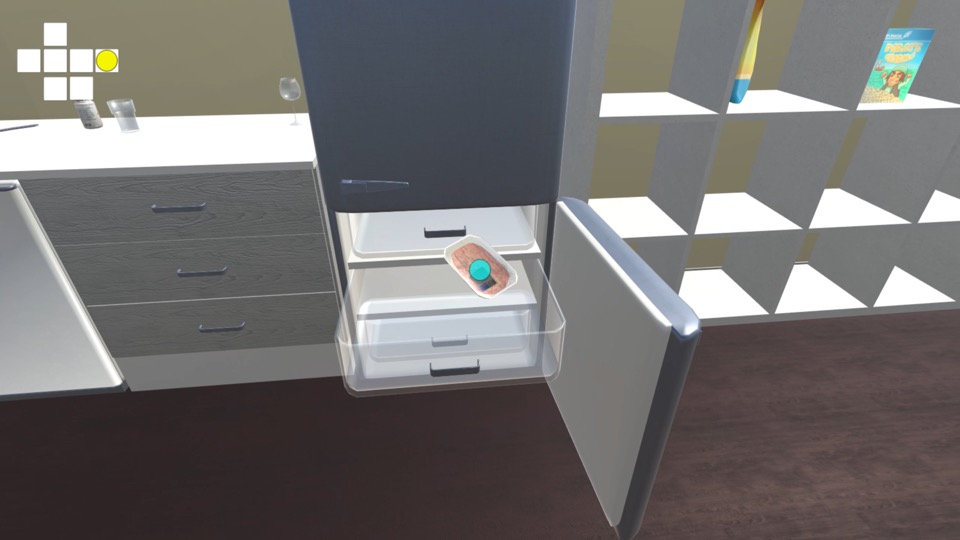} &
                  \tablefig{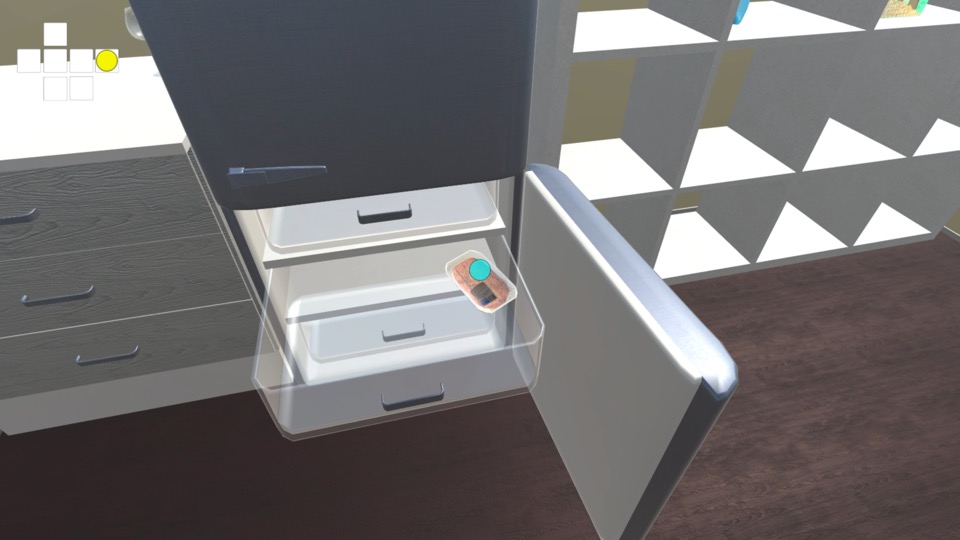} &
                  \tablefig{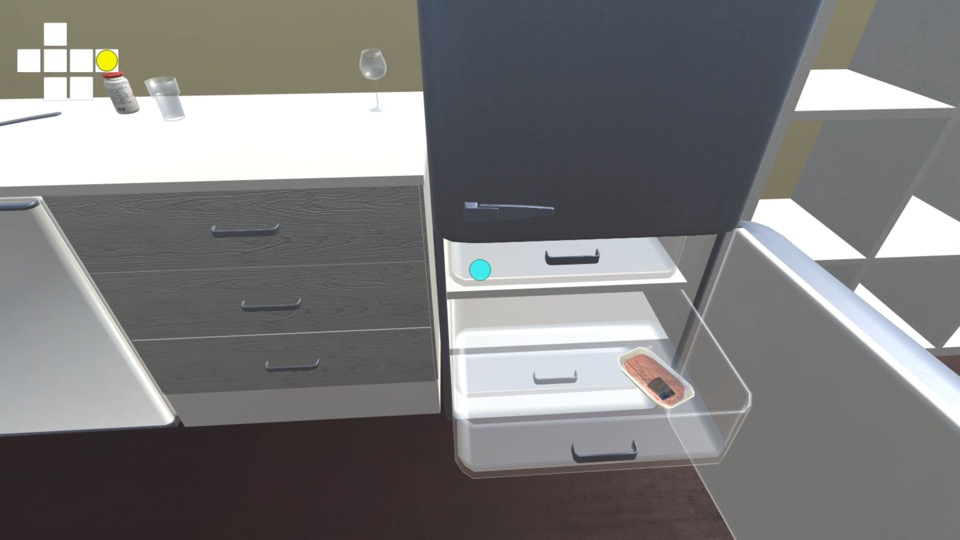} &
                  \tablefig{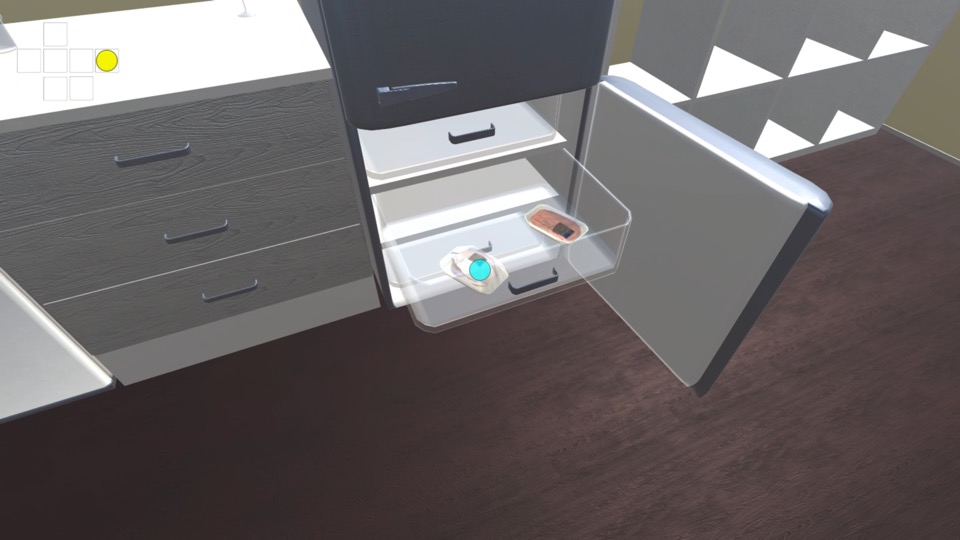} &
                  \tablefig{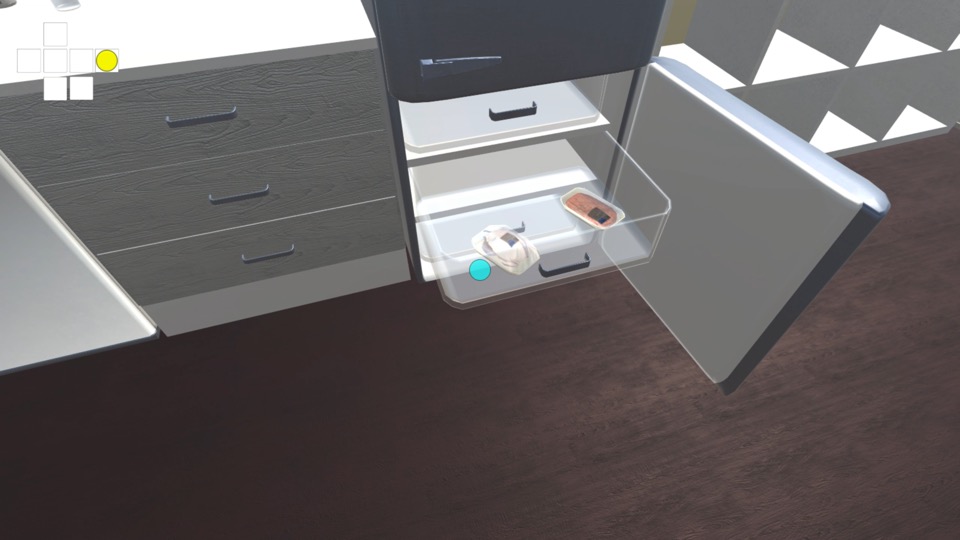} 
          \end{tabular}
          \vspace{5pt}
      		\end{minipage}}\\[1.5em]
            
     \fbox{\begin{minipage}{0.98\linewidth}
     \centering
     \vspace{5pt}
		 \begin{tabular}{cccccc}            
             \tablefig{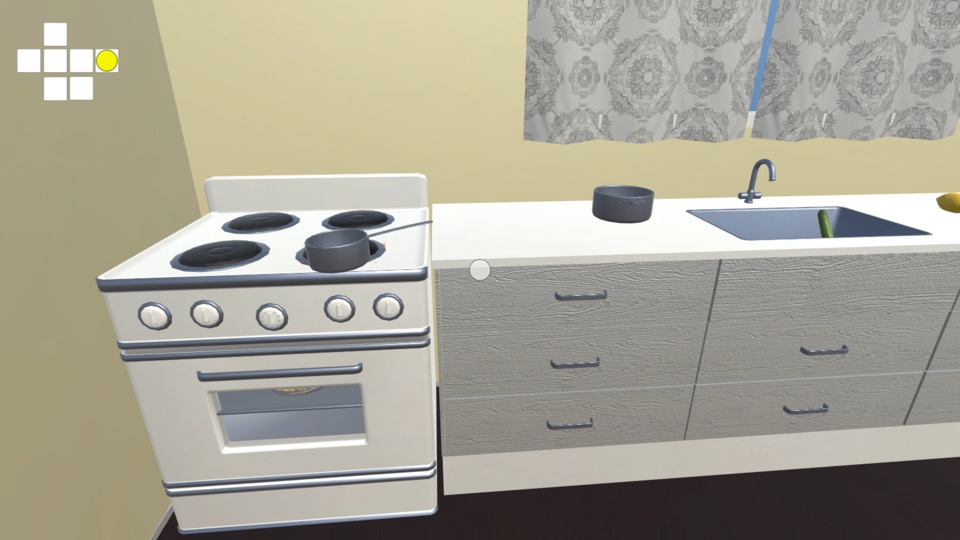} & 
             \tablefig{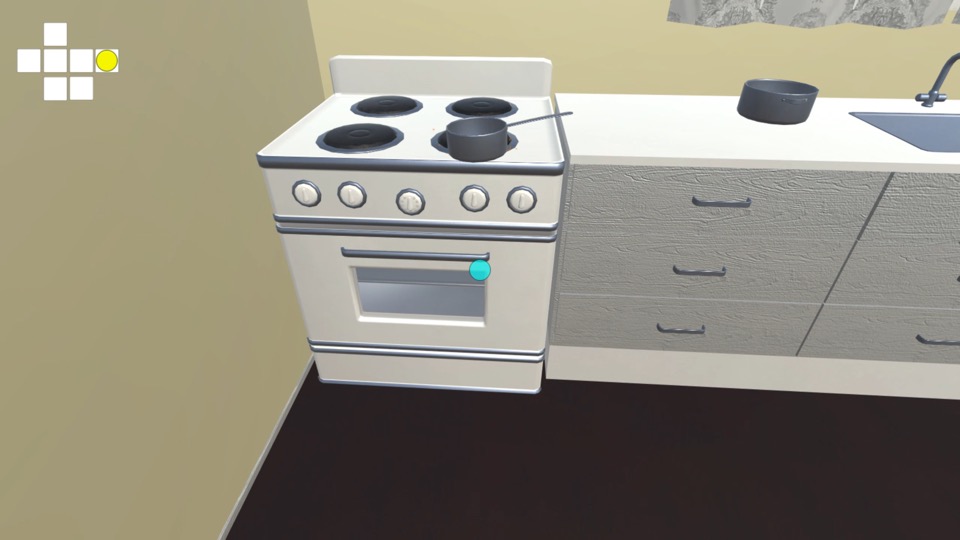} & 
             \tablefig{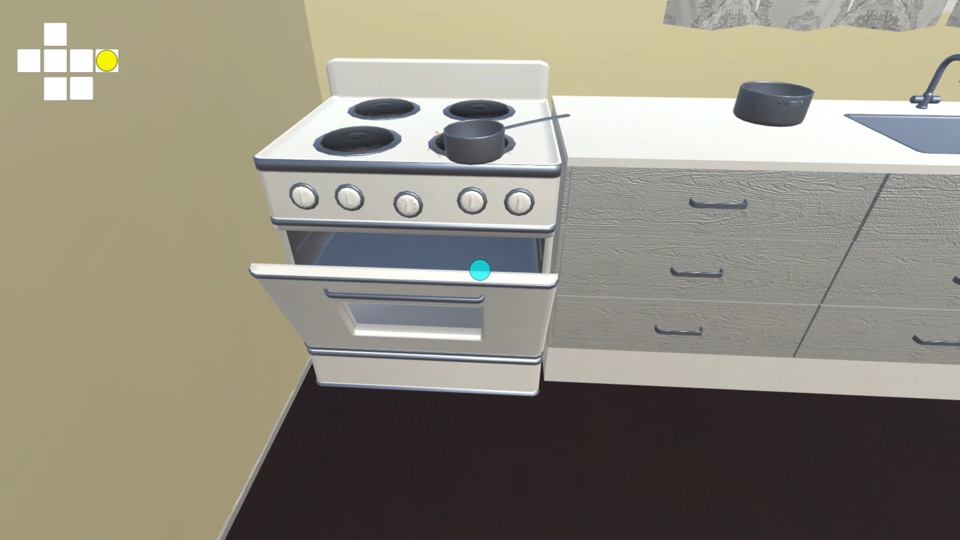} & 
             \tablefig{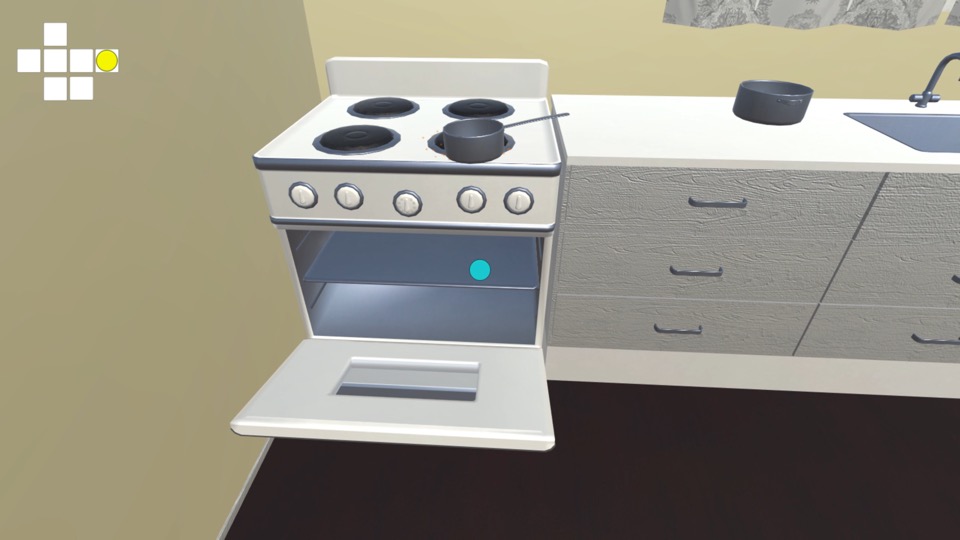} & 
             \tablefig{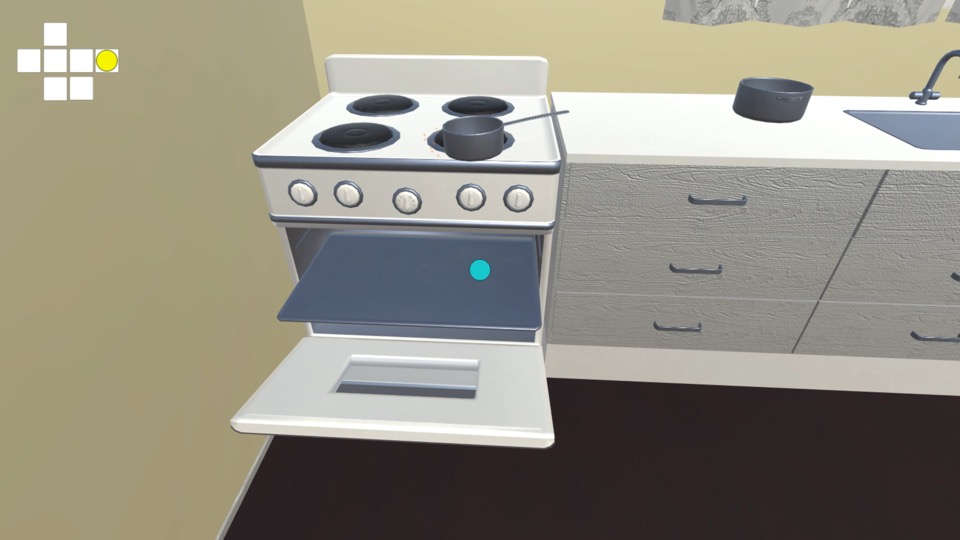} & 
             \tablefig{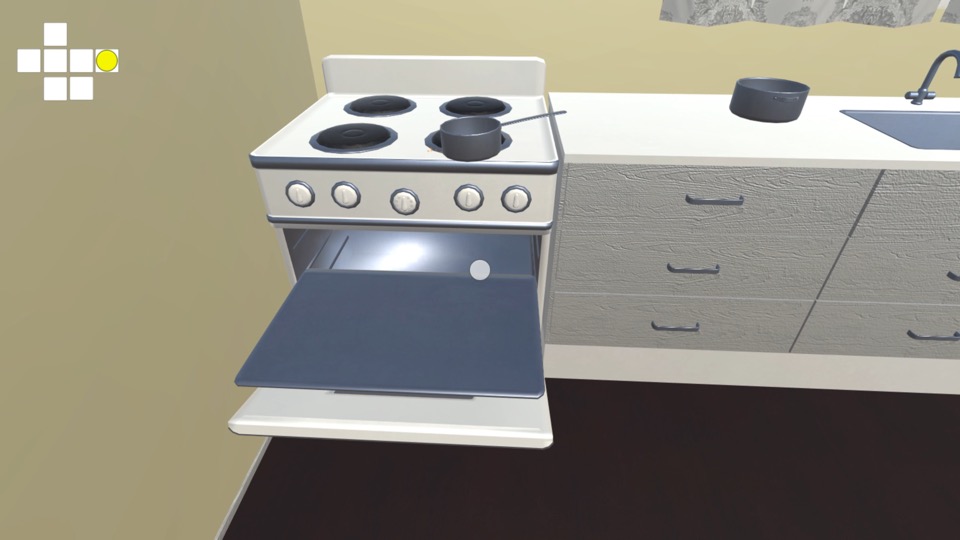} \\
             \tablefig{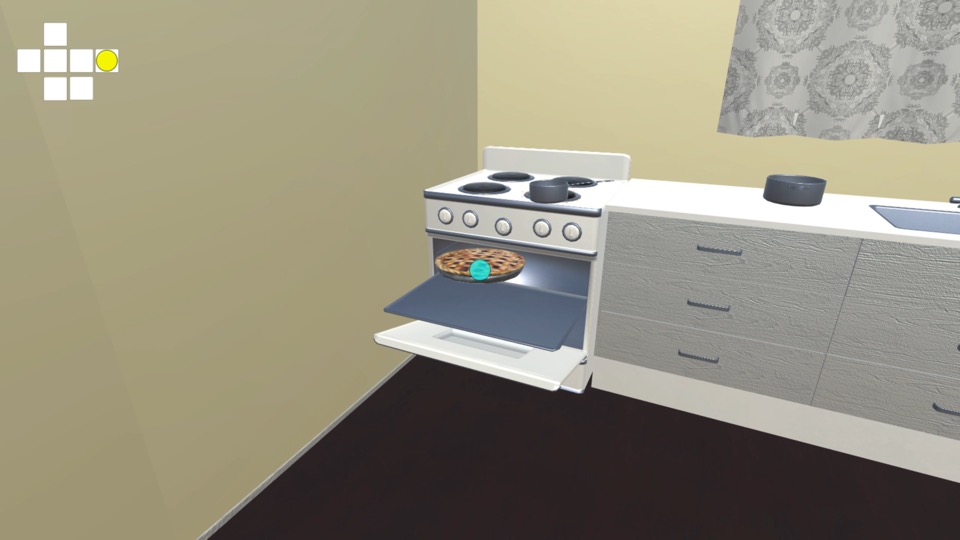} & 
             \tablefig{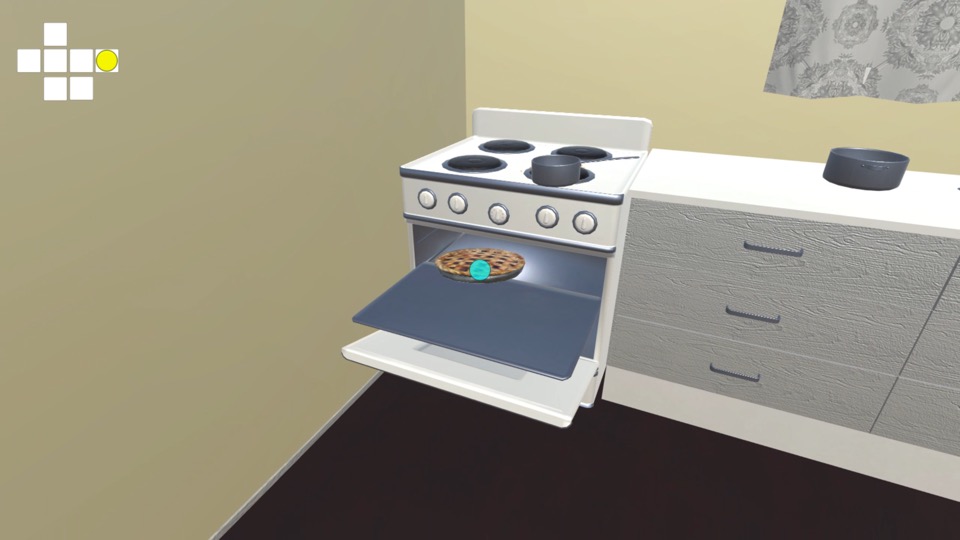} & 
             \tablefig{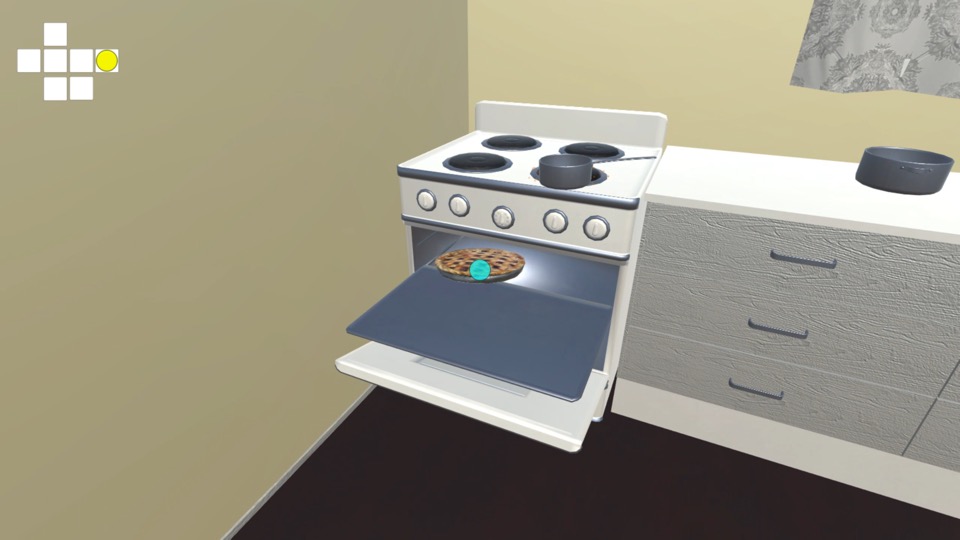} & 
             \tablefig{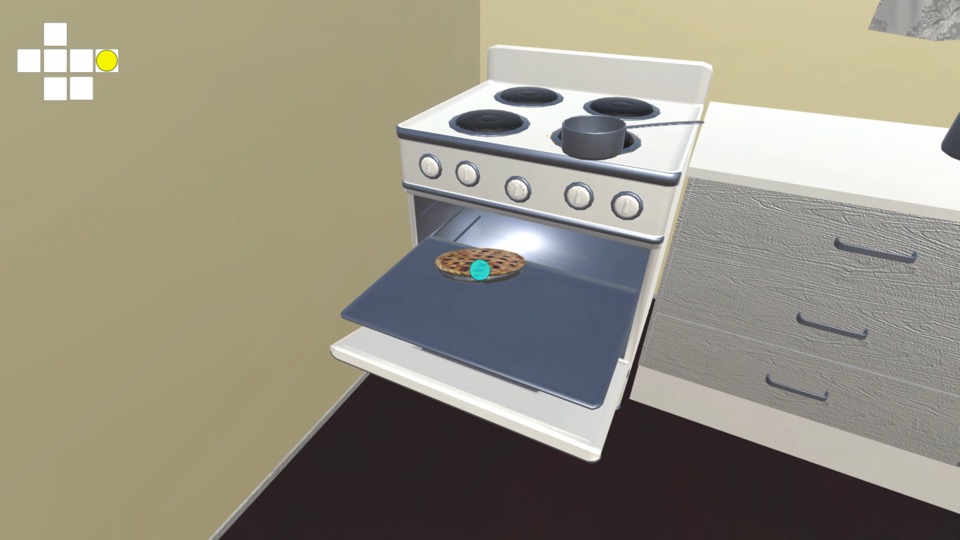} & 
             \tablefig{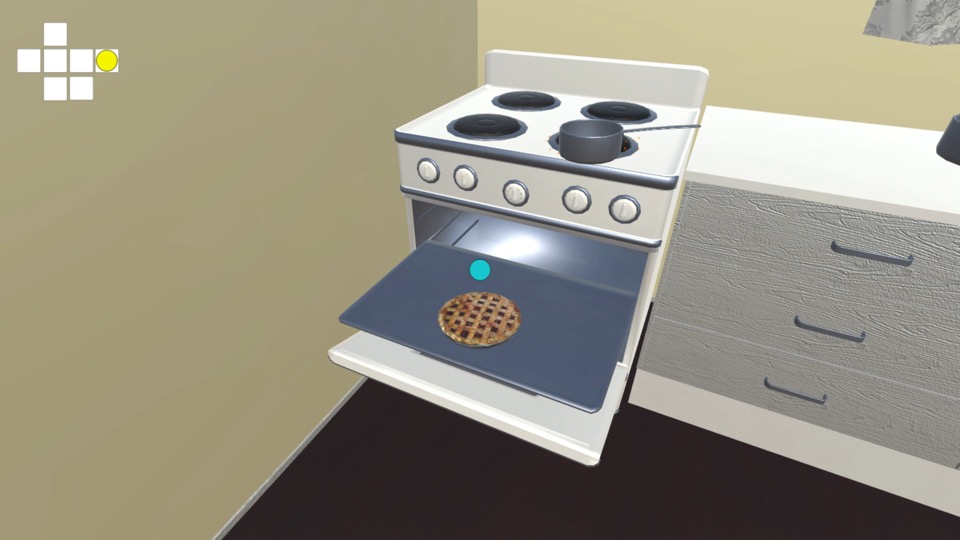} & 
             \tablefig{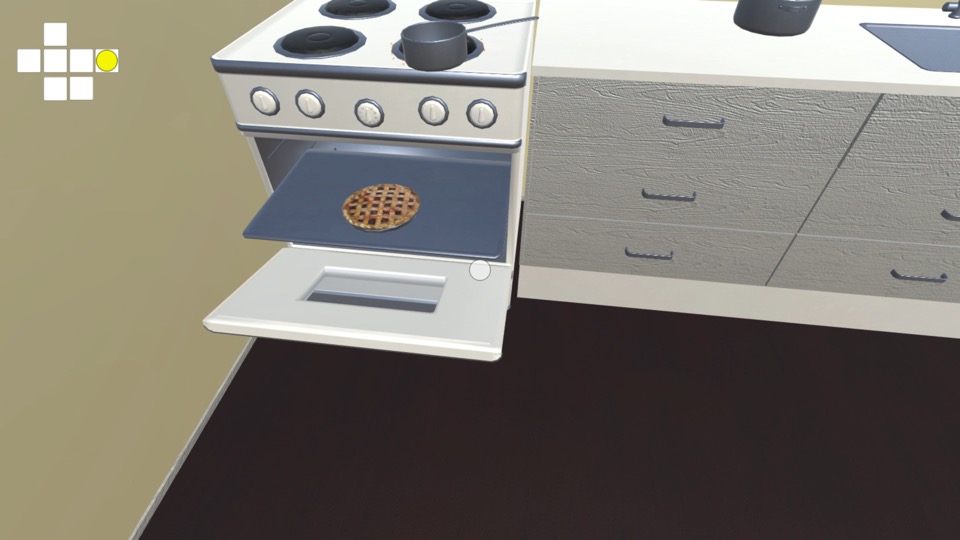}
        \end{tabular}
        \vspace{5pt}
        \end{minipage}}\\[1.5em]

		\caption{Sampled observations from three sequences of object manipulation, from top to bottom: loading the dishwasher, placing a food item in the freezer, and placing a pie in the oven.}
        \label{fig:continuous-control}
    \end{figure*}

The agent in \software{} observes the environment from a first-person perspective. At any given time, the agent observes only what is in front of it. 
The agent position is parameterized by two coordinates for its location and two coordinates for the orientation of its view. 
Changing the agent location is done in the direction of its orientation (i.e., first-person coordinate system). Whether the agent looks up or down does not influence location changes. 
All agent actions are continuous, but can be discretized to pre-defined precision by specifying the quantities of change for each step. 
Table~\ref{tbl-action} describes the agent actions.

\software{}  provides a rich testbed for language, exploration, planning, and modeling challenges. 
We design rooms to often include many objects of the same types. Instructions or questions that refer to a specific objet must then use spatial relations and object properties to identify the exact instance. 
For example, to pick up a specific towel in a bathroom, the agent is likely to be given an instruction such as \nlstring{pick up the yellow towel left of the sink}. 
In contrast, in an environment with a single object of each type, it would have been sufficient to ask to \nlstring{pick up the towel}. 
The ability to open and close containers also creates several interesting challenges. 
Given an instruction, such as \nlstring{put the glass from the cupboard on the table}, it is insufficient for an agent to simply align the word \nlstring{glass} to an observed object. Instead, it must resolve the noun phrase \nlstring{the cupboard} and the relation \nlstring{from} to understand it must look for a glass in a specific location. 
Simply resolving the target object (\nlstring{glass}) is insufficient. 
If multiple cupboards are available, the agent must also explore the different cupboards to find the one containing a glass. This requires both deciding on an exploration policy, and planning a complex sequence of actions. 
Finally, the agent perspective requires models that support access to previous observations or a representation of them (i.e., memory) to overcome the partial observability challenge.  

\section{Evaluation in CHALET}

Evaluating agent performance in CHALET is done by comparing the agent behavior to an annotated demonstration. A demonstration is a sequence of states and actions $\tau = \langle s_0, a_1, s_1, a_2, \cdots a_m, s_m \rangle$, where $s_i$ is state and $a_i$ is the action taken at that state. The start state of the demonstration is $s_0$ and the final state is $s_m$. 
A state $s_i$  contains information about the position, the orientation, and the interaction state of every object in the house, including the agent. 
We use two metrics for evaluating navigation errors and manipulation accurcy. 
Navigation error is the sum of euclidean distances in each room the agent must travel to reach the goal. For the room the agent is currently located in, we take the euclidean distance between the agent and the door to the next room. For each intermediate room, we take the euclidean distance between the door where the agent enters and the door to the following room. For the room containing the goal (i.e., the agent position in $s_m$), we measure the distance between the entry door to the goal. In each room, the distance measured is a straight line. 
We compute manipulation accuracy by extracting an ordered list of interaction actions from the annotated demonstration $\tau$. 
An interaction action may be picking an object, placing an object at a specific location, and changing the interaction state of an object (e.g., opening a drawer). All objects are uniquely identified. 
The manipulation accuracy is the F1-score computed for the list of actions extracted from the agent execution against the reference list of actions extracted from $\tau$.  
We consider placing an object within a radius of 1.0m  of the specified position in the same room as equivalent.

\section{Implementation Details}

\software{} is implemented in Unity 3D,\footnote{https://unity3d.com/}  a professional game development engine.\footnote{The community version of Unity, which was used to develop \software{} is publicly for education purposes.} 
The environment logic is written in the C\# scripting language, which supports high-level object-oriented programming constructs and is tightly integrated with the Unity engine. 
Using Unity provides several advantages. \software{} can be easily compiled for different platforms, including Linux, MacOS, Windows, Android, iOS, and WebGL. 
Unity also provides a built in physics engine, and  supports integration with augmented- and virtual-reality devices. 
Extending \software{} with new objects from the Unity Asset Store\footnote{https://www.assetstore.unity3d.com/en/} is trivial.

\software{} supports three modes of operation:

\begin{itemize}
\item {\tt standalone}: actions are provided using keyboard and mouse input. The generated trajectory is saved to a file. This model is used for crowdsourcing using a WebGL build. 
\item {\tt simulator}: actions are read from a saved file and executed in sequence. This mode allows replaying previously recorded trajectories, for example during crowdsourcing. 
\item {\tt client}: a separate process provides actions, and the framework returns the agent observations and information about the environment as required. Communication is done over sockets. This mode enables interaction with machine learning frameworks. 
\end{itemize}

\noindent
The framework provides a simple API to compute reward and feedback signals, as required for learning, and provide information about the environment, including the position and state of objects.
\software{} also provides programmatic generation of rich scenarios by adding, removing, and re-placing objects  during runtime without the use of Unity or re-loading the simulator. 
To enable this, each room is annotated with a set of surface locations where items may be placed. Placing an objects requires specifying its type and orientation, and the target surface and coordinates.

\section{Related Environments}

\begin{table*}
\centering
\begin{tabular}{|c|c|c|c|c|c|}
\hline
\textbf{Software} & \textbf{Type} & \textbf{Number of Environments} & \textbf{Navigation} & \textbf{Manipulation}\\
\hline
MINOS \cite{Savva:17} & Simulated & 45K houses + variations & Yes & No\\
House3D \cite{Wu:17} & Simulated & Thousand houses & Yes & No\\
AI2-Thor \cite{Zhu:16TargetdrivenVN,Gordon:17} & Simulated & 120 rooms  & Yes & Yes \\
Matterport3D \cite{Anderson:17}  & Real Images & 90  houses& Yes & No \\
HoME \cite{Brodeur:17} & Simulated & 45000 houses & Yes & Yes \\
DeepMind Lab \cite{beattie2016deepmind} & Simulated & Few (procedural) & Yes & No \\
\hline
\software  & Simulated & 58 rooms and 10 default houses & Yes & Yes\\
\hline
\end{tabular}
\caption{Comparison of \software{} with other 3D house simulators}
\label{tbl-comparison}
\end{table*}

Table~\ref{tbl-comparison} compares \software{} to existing simulators. 
\citet{Savva:17}, \citet{Wu:17}, and \citet{beattie2016deepmind} provide similar observations to \software{} in navigation-only environments. In contrast, \software{} emphasizes manipulation of both objects and the environment to support complex tasks. 
\citet{Anderson:17} use real images with a discrete state space for navigation. 
While \software{} includes 3D rendered environments, it provides  a continuous environment with a variety of actions. 
The most related environments to ours are HoME~\cite{Brodeur:17} and AI2-Thor~\cite{Gordon:17}, both provide 3D rendered houses with object manipulation. 
Unlike HoME, which only supports moving objects, \software{} enables toggling the state of objects and changing the environment by modifying containers. 
In contrast to Thor, \software{} supports moving between rooms in complete houses, while the current version of Thor supports a single room.

There is also significant work on using simulators for other domains. 
Atari~\cite{Mnih:13atari}, OpenAI Gym~\cite{Brockman:openaigym}, Project Malmo~\cite{johnson2016malmo}, Minecraft~\cite{Oh:16rl-minecraft}, Gazebo~\cite{Zamora:16gazebo}, Viz Doom~\cite{Kempka16} are commonly used for testing reinforcement learning algorithms. 
Simulators have also been used to evaluate natural language instruction following~\cite{Misra:15highlevel,Bisk:2016:NAACL,Misra:17instructions,Janner:17spatialrep,Hermann:17} and   question answering~\cite{Gordon:17,embodiedqa,Johnson:16clevr,Suhr:17visual-reason}. 
The manipulation features and partial observability challenges of \software{} provide a more realistic testbed for studying language, including for instruction following, visual reasoning, and question answering.

\section*{Acknowledgments}

This work was supported by NSF under Grant No. 1750499, the generosity of Eric and Wendy Schmidt by recommendation of the Schmidt Futures program, and the Women in Technology and Entrepreneurship in New York (WiTNY) initiative. 
We also thank Porrith Suong for help with Unity3D development.

\bibliographystyle{plainnat}
\bibliography{references}

\begin{thebibliography}{22}
\providecommand{\natexlab}[1]{#1}
\providecommand{\url}[1]{\texttt{#1}}
\expandafter\ifx\csname urlstyle\endcsname\relax
  \providecommand{\doi}[1]{doi: #1}\else
  \providecommand{\doi}{doi: \begingroup \urlstyle{rm}\Url}\fi

\bibitem[Anderson et~al.(2017)Anderson, Wu, Teney, Bruce, Johnson,
  S{\"{u}}nderhauf, Reid, Gould, and van~den Hengel]{Anderson:17}
Peter Anderson, Qi~Wu, Damien Teney, Jake Bruce, Mark Johnson, Niko
  S{\"{u}}nderhauf, Ian~D. Reid, Stephen Gould, and Anton van~den Hengel.
\newblock Vision-and-language navigation: Interpreting visually-grounded
  navigation instructions in real environments.
\newblock \emph{arXiv preprint arXiv:1711.07280}, 2017.

\bibitem[Beattie et~al.(2016)Beattie, Leibo, Teplyashin, Ward, Wainwright,
  K{\"u}ttler, Lefrancq, Green, Vald{\'e}s, Sadik, et~al.]{beattie2016deepmind}
Charles Beattie, Joel~Z Leibo, Denis Teplyashin, Tom Ward, Marcus Wainwright,
  Heinrich K{\"u}ttler, Andrew Lefrancq, Simon Green, V{\'\i}ctor Vald{\'e}s,
  Amir Sadik, et~al.
\newblock Deepmind lab.
\newblock \emph{arXiv preprint arXiv:1612.03801}, 2016.

\bibitem[Bisk et~al.(2016)Bisk, Yuret, and Marcu]{Bisk:2016:NAACL}
Yonatan Bisk, Deniz Yuret, and Daniel Marcu.
\newblock Natural language communication with robots.
\newblock In \emph{Proceedings of the 15th Annual Conference of the North
  American Chapter of the Association for Computational Linguistics}, pages
  751--761, San Diego, CA, June 2016.

\bibitem[Brockman et~al.(2016)Brockman, Cheung, Pettersson, Schneider,
  Schulman, Tang, and Zaremba]{Brockman:openaigym}
Greg Brockman, Vicki Cheung, Ludwig Pettersson, Jonas Schneider, John Schulman,
  Jie Tang, and Wojciech Zaremba.
\newblock Openai gym, 2016.

\bibitem[Brodeur et~al.(2017)Brodeur, Perez, Anand, Golemo, Celotti, Strub,
  Rouat, Larochelle, and Courville]{Brodeur:17}
Simon Brodeur, Ethan Perez, Ankesh Anand, Florian Golemo, Luca Celotti, Florian
  Strub, Jean Rouat, Hugo Larochelle, and Aaron~C. Courville.
\newblock {HoME: a Household Multimodal Environment}.
\newblock \emph{arXiv preprint arXiv:1711.11017}, 2017.

\bibitem[Das et~al.(2017)Das, Datta, Gkioxari, Lee, Parikh, and
  Batra]{embodiedqa}
Abhishek Das, Samyak Datta, Georgia Gkioxari, Stefan Lee, Devi Parikh, and
  Dhruv Batra.
\newblock {E}mbodied {Q}uestion {A}nswering.
\newblock \emph{arXiv preprint arXiv:1711.11543}, 2017.

\bibitem[Gordon et~al.(2017)Gordon, Kembhavi, Rastegari, Redmon, Fox, and
  Farhadi]{Gordon:17}
Daniel Gordon, Aniruddha Kembhavi, Mohammad Rastegari, Joseph Redmon, Dieter
  Fox, and Ali Farhadi.
\newblock {IQA: Visual Question Answering in Interactive Environments}.
\newblock \emph{arXiv preprint arXiv:1712.03316}, 2017.

\bibitem[Hermann et~al.(2017)Hermann, Hill, Green, Wang, Faulkner, Soyer,
  Szepesvari, Czarnecki, Jaderberg, Teplyashin, Wainwright, Apps, Hassabis, and
  Blunsom]{Hermann:17}
Karl~Moritz Hermann, Felix Hill, Simon Green, Fumin Wang, Ryan Faulkner, Hubert
  Soyer, David Szepesvari, Wojciech~Marian Czarnecki, Max Jaderberg, Denis
  Teplyashin, Marcus Wainwright, Chris Apps, Demis Hassabis, and Phil Blunsom.
\newblock {Grounded Language Learning in a Simulated 3D World}.
\newblock \emph{arXiv preprint arXiv:1706.06551}, 2017.

\bibitem[J{\"a}nner et~al.(2017)J{\"a}nner, Narasimhan, and
  Barzilay]{Janner:17spatialrep}
Michaela J{\"a}nner, Karthik Narasimhan, and Regina Barzilay.
\newblock Representation learning for grounded spatial reasoning.
\newblock \emph{CoRR}, abs/1707.03938, 2017.

\bibitem[Johnson et~al.(2016{\natexlab{a}})Johnson, Hariharan, van~der Maaten,
  Fei-Fei, Zitnick, and Girshick]{Johnson:16clevr}
Justin Johnson, Bharath Hariharan, Laurens van~der Maaten, Li~Fei-Fei,
  C.~Lawrence Zitnick, and Ross~B. Girshick.
\newblock {CLEVR}: A diagnostic dataset for compositional language and
  elementary visual reasoning.
\newblock \emph{CoRR}, abs/1612.06890, 2016{\natexlab{a}}.

\bibitem[Johnson et~al.(2016{\natexlab{b}})Johnson, Hofmann, Hutton, and
  Bignell]{johnson2016malmo}
Matthew Johnson, Katja Hofmann, Tim Hutton, and David Bignell.
\newblock The malmo platform for artificial intelligence experimentation.
\newblock In \emph{International Joint Conferences on Artificial Intelligence},
  pages 4246--4247, 2016{\natexlab{b}}.

\bibitem[Kempka et~al.(2016)Kempka, Wydmuch, Runc, Toczek, and
  Jaskowski]{Kempka16}
Michal Kempka, Marek Wydmuch, Grzegorz Runc, Jakub Toczek, and Wojciech
  Jaskowski.
\newblock Vizdoom: {A} doom-based {AI} research platform for visual
  reinforcement learning.
\newblock \emph{arXiv preprint arXiv:1605.02097}, 2016.

\bibitem[Misra et~al.(2017)Misra, Langford, and Artzi]{Misra:17instructions}
Dipendra Misra, John Langford, and Yoav Artzi.
\newblock Mapping instructions and visual observations to actions with
  reinforcement learning.
\newblock In \emph{Proceedings of the 2017 Conference on Empirical Methods in
  Natural Language Processing (EMNLP)}. Association for Computational
  Linguistics, 2017.

\bibitem[Misra et~al.(2015)Misra, Tao, Liang, and Saxena]{Misra:15highlevel}
Kumar~Dipendra Misra, Kejia Tao, Percy Liang, and Ashutosh Saxena.
\newblock Environment-driven lexicon induction for high-level instructions.
\newblock In \emph{Proceedings of the 53rd Annual Meeting of the Association
  for Computational Linguistics and the 7th International Joint Conference on
  Natural Language Processing (Volume 1: Long Papers)}, 2015.
\newblock \doi{10.3115/v1/P15-1096}.

\bibitem[Mnih et~al.(2013)Mnih, Kavukcuoglu, Silver, Graves, Antonoglou,
  Wierstra, and Riedmiller]{Mnih:13atari}
Volodymyr Mnih, Koray Kavukcuoglu, David Silver, Alex Graves, Ioannis
  Antonoglou, Daan Wierstra, and Martin~A. Riedmiller.
\newblock Playing atari with deep reinforcement learning.
\newblock In \emph{Advances in Neural Information Processing Systems}, 2013.

\bibitem[Oh et~al.(2016)Oh, Chockalingam, Singh, and Lee]{Oh:16rl-minecraft}
Junhyuk Oh, Valliappa Chockalingam, Satinder~P. Singh, and Honglak Lee.
\newblock Control of memory, active perception, and action in minecraft.
\newblock In \emph{Proceedings of the International Conference on Machine
  Learning}, 2016.

\bibitem[Salimans et~al.(2017)Salimans, Ho, Chen, and
  Sutskever]{salimans2017evolution}
Tim Salimans, Jonathan Ho, Xi~Chen, and Ilya Sutskever.
\newblock Evolution strategies as a scalable alternative to reinforcement
  learning.
\newblock \emph{arXiv preprint arXiv:1703.03864}, 2017.

\bibitem[Savva et~al.(2017)Savva, Chang, Dosovitskiy, Funkhouser, and
  Koltun]{Savva:17}
Manolis Savva, Angel~X. Chang, Alexey Dosovitskiy, Thomas Funkhouser, and
  Vladlen Koltun.
\newblock {MINOS: Multimodal Indoor Simulator for Navigation in Complex
  Environments}.
\newblock \emph{arXiv preprint arXiv:1712.03931v1}, 2017.

\bibitem[Suhr et~al.(2017)Suhr, Lewis, Yeh, and Artzi]{Suhr:17visual-reason}
Alane Suhr, Mike Lewis, James Yeh, and Yoav Artzi.
\newblock A corpus of natural language for visual reasoning.
\newblock In \emph{Proceedings of the 55th Annual Meeting of the Association
  for Computational Linguistics}. Association for Computational Linguistics,
  2017.

\bibitem[Wu et~al.(2017)Wu, Wu, Gkioxari, and Tian]{Wu:17}
Yi~Wu, Yuxin Wu, Georgia Gkioxari, and Yuandong Tian.
\newblock {Building Generalizable Agents with a Realistic and Rich 3D
  Environment}.
\newblock \emph{arXiv preprint arXiv:1801.02209v1}, 2017.

\bibitem[Zamora et~al.(2016)Zamora, Lopez, Vilches, and
  Cordero]{Zamora:16gazebo}
Iker Zamora, Nestor~Gonzalez Lopez, Victor~Mayoral Vilches, and
  Alejandro~Hernandez Cordero.
\newblock Extending the openai gym for robotics: a toolkit for reinforcement
  learning using ros and gazebo.
\newblock \emph{arXiv preprint arXiv:1608.05742}, 2016.

\bibitem[Zhu et~al.(2017)Zhu, Mottaghi, Kolve, Lim, Gupta, Fei-Fei, and
  Farhadi]{Zhu:16TargetdrivenVN}
Yuke Zhu, Roozbeh Mottaghi, Eric Kolve, Joseph~J. Lim, Abhinav Gupta,
  Li~Fei-Fei, and Ali Farhadi.
\newblock Target-driven visual navigation in indoor scenes using deep
  reinforcement learning.
\newblock In \emph{IEEE International Conference on Robotics and Automation},
  2017.

\end{thebibliography}

\end{document}